\documentclass{llncs}
\usepackage{llncsdoc}
\usepackage{graphicx}
\usepackage{float}
\usepackage{multirow}
\usepackage{color}
\usepackage{bbold}
\usepackage{caption}

\begin{document}
\title{Explaining Radiological Emphysema Subtypes with Unsupervised Texture Prototypes: \\
MESA COPD Study}

\author{Jie Yang \inst{1} \and Elsa D. Angelini\inst{1,2} \and Benjamin M. Smith \inst{3,4} \and  \\
John H.M. Austin \inst{2} \and Eric A. Hoffman \inst{5,6} \and David A. Bluemke\inst{7} \and \\
R. Graham Barr\inst{3,8} \and Andrew F. Laine\inst{1}}

\institute{Dept. of Biomedical Engineering, Columbia University, New York, NY, USA
\and Dept. of Radiology, Columbia University Medical Center, New York, NY, USA
\and Dept. of Medicine, Columbia University Medical Center, New York, NY, USA
\and Dept. of Medicine, McGill University Health Center, Montreal, QC, Canada
\and Dept. of Radiology, University of Iowa, Iowa City, IA, USA
\and Dept. of Biomedical Engineering, University of Iowa, Iowa City, IA, USA
\and Radiology and Imaging Sciences, National Institutes of Health, Bethesda, MD
\and Dept. of Epidemiology, Columbia University Medical Center, New York, NY, USA}

\maketitle
\begin{abstract}

Pulmonary emphysema is traditionally subcategorized into three subtypes, which have distinct radiological appearances on computed tomography (CT) and can help with the diagnosis of chronic obstructive pulmonary disease (COPD). Automated texture-based quantification of emphysema subtypes has been successfully implemented via supervised learning of these three emphysema subtypes. In this work, we demonstrate that unsupervised learning on a large heterogeneous database of CT scans can generate texture prototypes that are visually homogeneous and distinct, reproducible across subjects, and capable of predicting accurately the three standard radiological subtypes. These texture prototypes enable automated labeling of lung volumes, and open the way to new interpretations of lung CT scans with finer subtyping of emphysema.  

\end{abstract}

\section{Introduction}

Chronic obstructive pulmonary disease (COPD), characterized by limitation of airflow, is a leading cause of morbidity and mortality \cite{gold}. Pulmonary emphysema, defined by a loss of lung tissue in the absence of fibrosis, overlaps considerably with COPD. 

Pulmonary emphysema is traditionally subcategorized into three standard subtypes, which were initially defined at autopsy, and can be visually assessed on computed tomography (CT), according to the following definitions \cite{ben}: \emph{centrilobular emphysema} (CLE), defined as focal regions of low attenuation surrounded by normal lung attenuation; \emph{panlobular emphysema} (PLE), defined as diffuse regions of low attenuation involving entire secondary pulmonary lobules; and \emph{paraseptal emphysema} (PSE), defined as regions of low attenuation adjacent to visceral pleura (including fissures). Given that these subtypes are associated with distinct risk factors and clinical manifestations \cite{riskfactor}\cite{riskfactor2}, they are therefore likely to represent different diseases and can help with the diagnosis of COPD. 

Radiologists' interpretation of standard subtypes is labor-intensive, and has modest inter-rater agreements \cite{ben,barr}. Automated texture-based analysis of emphysema offers the potential of automated COPD diagnosis and catalyzing research (e.g. discovering emphysema subtypes), and is receiving increasing interest \cite{texture2006,lbp2,texton2010,R1,review}. However, most existing approaches are limited to supervised emphysema subtype classification using manually annotated scans in local regions of interest (ROIs), which are very costly and time-consuming to obtain. Furthermore, it is unclear if the supervised classifiers generalize to other datasets with varying in-plane resolutions and scanner types.

A recent clinical study \cite{ben} demonstrated the reliability and clinical significance of global (rather than local) labeling of lung volumes using the three standard subtypes. Global labeling generates weakly labeled data that was used for the classification of COPD subjects with multiple instance learning (MIL) \cite{MIL}. However, MIL has only been demonstrated so far for binary labeling of emphysema versus normal tissue, rather than to distinguish the three subtypes, and can generate unreliable local ROI labeling. 

In this work, we present a novel framework to discover unsupervised fine-grained prototypes that go beyond but still have the power of encoding the three standard emphysema subtypes. Our method clusters local ROIs of lung volumes into texture prototypes in an unsupervised manner, and builds signatures of lung volumes with texture prototype histograms. The extent of standard emphysema subtypes can be predicted from these prototype histograms with a constrained multivariate regression on global labels. To our knowledge, this is the first study whereby texture-based predictions are used to globally characterize the standard emphysema subtypes.

Three types of texture features were tested, extracted from 3D or 2D local ROIs, to generate the emphysema prototypes: 1) frequency histograms of textons (called texton-based features), used in \cite{texton2010}\cite{R1}; 2) soft histograms of intensities and difference of Gaussian (DoG) responses (called DOG2 features), used in \cite{ltp}; and 3) joint histograms of local binary patterns (LBP) and intensities (called LBP2 features), used in \cite{lbp2}. 

\section{Method}

\subsection{Framework Overview}

\begin{figure}[htbp]
\includegraphics[width=12cm]{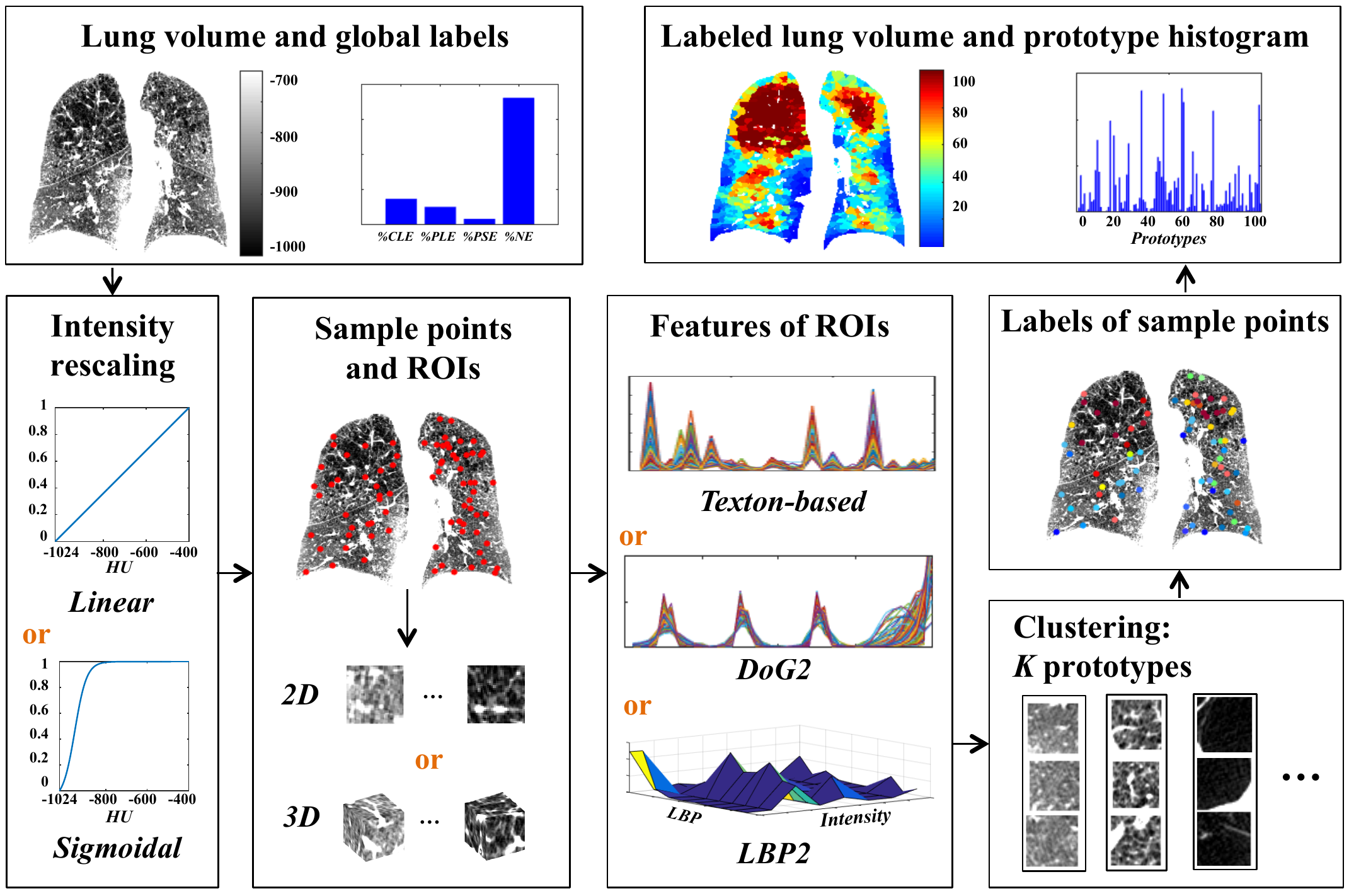}
\caption[ ] {Illustration of the pipeline for generating texture prototype histograms.}
\label{fig:framework}
\end{figure}

Our framework is divided into a learning stage in an unsupervised sense, and a prediction stage of radiological emphysema subtypes using globally annotated data. The intensity of lung voxels, inside lung masks generated using the APOLLO\textsuperscript{\textregistered} software (VIDA Diagnostics, Coralville, Iowa), are rescaled from [-1024, -400] HU to [0, 1] via either linear or sigmoidal mapping in pre-processing.

In the learning stage, texture prototypes are learned and prototype histograms $H_p$ are built for each training lung volume. 

Fig. \ref{fig:framework} illustrates the pipeline for generating prototype histograms.  Sample points are randomly extracted uniformly within the lung volumes. 2D or 3D neighborhoods of sampled points are used as local ROIs, with a size of $25\mathrm{mm}^2$ or $25\mathrm{mm}^3$, approximating the diameter of secondary pulmonary lobules. Our target number of sample points per scan is $N=\mathrm{(lung\ volume)}/25\mathrm{mm}^3$. Since we discard ROIs with more that 50\% of non-lung field, we adjust the sample ratio $\alpha$ so that $\alpha \cdot N - N_{discarded}=N$. The value $\alpha=5$ is suitable for the population of scans, leading to an average of 1,512 sample points per CT scan. ROIs are characterized with texture features (texton-based, DOG2 or LBP2), and are clustered into $K$ texture prototypes in an unsupervised manner. For interpretation, prototypes are ordered according to the average intensity value of training ROIs belonging to each prototype. Each sample point is labeled with the prototype centroid most similar to its ROI (i.e. with least distance in feature space). Finally, other voxels within the lung volumes are labeled by assigning the prototype label of the nearest sample point. 

In the prediction stage, sample points and ROIs are extracted from test lung volumes and ROI texture features are generated. ROIs are labeled by assigning the most similar prototype centroid. Prototype histograms are then generated for test lung volumes following the same procedure as in the training stage. 

To evaluate our texture prototypes, we regressed their occurrence against global emphysema labels in \cite{ben} on training scans, with a constrained multivariate model. Global labels $H_g$ encode the extent of standard emphysema subtypes referred to as $\%\mathrm{CLE}$, $\%\mathrm{PLE}$, $\%\mathrm{PSE}$. The residual, denoted $\%\mathrm{NE}$, corresponds to tissue without emphysema (but maybe with some lung diseases).

In the following sections, we detail the texture features, the unsupervised learning of prototypes and the regression model.

\subsection{Texture Features}

\subsubsection{Texton-based Features:}
Texton-based features characterize ROIs with the help of a texton codebook. The texton codebook is formed by the cluster centers of intensity values (after linear mapping) from small-sized local patches (here 3 voxels in each dimension) randomly extracted from ROIs in the training set. Clustering is performed with \textit{K}-means. By projecting all small-sized patches onto the codebook, the texton-based feature of the ROI is the normalized histogram of texton frequencies. Targeting 4 classes and 10 textons per class \cite{texton2010}, the feature vector length is set to 40, using a codebook with 40 textons. 

Note that our texton prototype histogram uses the bag-of-words (BoW)\cite{bow} model on two scales: 1) building of ROI-level texture features based on a texton dictionary; 2) building subject-level lung CT signatures based on texture prototypes. To our knowledge, BoW has not been exploited for subject-level signatures before.

\subsubsection{DOG2 Features:}
The DOG2 feature of a ROI is a concatenation of four normalized soft histograms: one intensity histogram, and three histograms of DoG responses at three octaves. Using 10 bins for each histogram, following the setting in \cite{ltp}, leads to a feature vector of length 40.

Intensity values in CT scans encode X-ray attenuations in Hounsfield units (HU) and their range is very large. To focus the texture learning process on the intensity range of interest (lung parenchyma and air), a sigmoid function is used, as in \cite{ltp}, to map values to the interval $[0 \ 1]$ with the highest contrast assigned to the range $[-1000 -900]$ HU where textural characteristics due to emphysema are presumed to be present.

\subsubsection{LBP2 Features:}
The LBP2 feature of a ROI is the joint histogram of LBP codes and intensity values (after linear mapping) of each voxel within the ROI. The LBP codes are obtained by thresholding samples in a local neighborhood around center voxel $x$. Formally:

\begin{equation}
LBP(x;R,P)= \sum_{p=0}^{P-1}H(I(x_p)-I(x))2^p 
\end{equation}
where $I(x)$ is the intensity of center voxel, $x_p$ are $P$ voxels sampled around $x$ at a given radial distance $R$, and $H(\cdot)$ is the Heaviside function. Rotational invariance is achieved by rotating the radial sampling until the lowest possible $LBP(x;R,P)$ value is found. We use 10 uniform rotational invariant LBP codes with R=1 and P=8, and 4 bins for the intensity histogram to match with other feature length, making the total feature length also 40 ($4\times10$).

\subsection{Prototype Clustering}
The number of prototypes $K$ should be large enough to handle the diversity of textures encountered in the lung volumes (i.e. good intra-prototype homogeneity), but small enough to avoid redundancy (i.e. good inter-prototype differences). Our strategy is to first select an empirically large number $K$ so as to generate homogenous prototypes and then trim the set to a smaller number of sufficient prototypes (number likely different for different texture features) according to a dedicated metric. We choose \textit{K}-means for the clustering task because of its efficiency at dealing with a large number of ROIs over scans. 

To trim the number of prototypes, instead of testing smaller $K$ values  with \textit{K}-means, which tends to decrease all intra-cluster homogeneity, we propose to merge prototypes iteratively according to their inter-prototype distance and spatial co-occurrence. 

The inter-prototype distance is measured by averaging the $\chi^2$ distance (common for histogram-based features) between each pair in feature space. The spatial co-occurrence of two prototypes $i$ and $j$ ($i\neq j$) is measured as:

\begin{equation}
S(i,j)=\frac{q(i,j)+q(j,i)}{\sum_{k=1}^{K}q(i,k)+\sum_{k=1}^{K}q(j,k)} 
\end{equation}
where $q(i,j)$ is the frequency of prototypes $i$ and $j$ appearing together in a pre-defined small neighborhood (here 10 voxels in each dimension).

At each iteration of the pruning process, each pair of prototypes is given a rank $R_{i,j}^{f}$ in inter-prototype distance (smallest ranks first), and a rank $R_{i,j}^{S}$ in spatial similarity (largest ranks first). The pair of prototypes to merge is the first one according to the rank: $R_{i,j}=R_{i,j}^{f}+R_{i,j}^{S}$.

\subsection{Constrained Multivariate Regression}
The probability of voxel $x$ belonging to a lung tissue class can be modeled as:

\begin{equation}
P\left(L(x) = C_i\right)=\sum_{k=1}^{K} P(L(x)=C_i|F(x)=p_k)P(F(x)=p_k)
\end{equation}
where $L(x)$ is the label of voxel $x$ as $C_i \in \{\mathrm{CLE, PLE, PSE, NE}\}$, and $F(x)$ is the voxel prototype label $p_k$ with $k \in 1,...,K$.
If prototypes are homogeneous, $P(L(x)=C_i|F(x)=p_k)$ can be assumed to be consistent throughout ROIs and subjects. We therefore infer the relation as:

\begin{equation}
Y_{N\times 4}=X_{N\times K}A_{K\times 4}
\end{equation}
where $N$ is the number of training scans. Each row in $Y$ is the global label $H_g=[P(L(x)=\mathrm{CLE}), P(L(x)=\mathrm{PSE}), P(L(x)=\mathrm{PLE}), P(L(x)=\mathrm{NE})]$ for one scan, each row in $X$ is the prototype histogram $H_p=[P(F(x)=p_1), ..., P(F(x)=p_K)]$ for the same scan, and $A$ is the matrix of regression coefficients with $A_{k,i}=P(L(x)=C_i|F(x)=p_k)$, $i = 1,...,4$ and $k=1,...,K$. We propose to learn $A$ with the following constrained multivariate regression model:

\begin{equation}
\mathrm{argmin}_A\|X_{train}A-Y_{train}\|_2,\ \mathrm{subject}\ \mathrm{to}\ 0<A_{k,i}<1\ \mathrm{and}\ \sum_{i=1}^{4}A_{k,i}=1
\end{equation}

\section{Results and Discussions}

\subsection{Data}
The dataset includes 321 full-lung CT scans from the Multi-Ethnic Study of Atherosclerosis (MESA) COPD Study \cite{ben}, among which 4 scans are discarded due to excessive motion artifact or incomplete lung field of view. All CT scans were acquired at full inspiration with either a Siemens 64-slice scanner or a GE 64-slice scanner, and reconstructed using B35/Standard kernels with axial resolutions within the range [0.58, 0.88]mm, and 0.625mm slice thickness. All scans were acquired at 120 kVp, 0.5 seconds, with milliamperes (mA) set by body mass index following the SPIROMICS protocol \cite{spiromics}.

Global labels of standard emphysema subtypes are available for each scan, corresponding to the average of visually assessed scores by four experienced radiologists \cite{ben}. Inter-rater intraclass correlations, evaluated on 40 random scans, are reported in Fig. \ref{fig:proto-pruning}. The clinically-evaluated prevalence of emphysema in this dataset is 27$\%$, with 14$\%$ CLE-predominance, 9$\%$ PSE-predominance, and 4$\%$ PLE-predominance. 

\subsection{Quality of Predictions}

\begin{figure}[t]
\centering
\includegraphics[width=12cm]{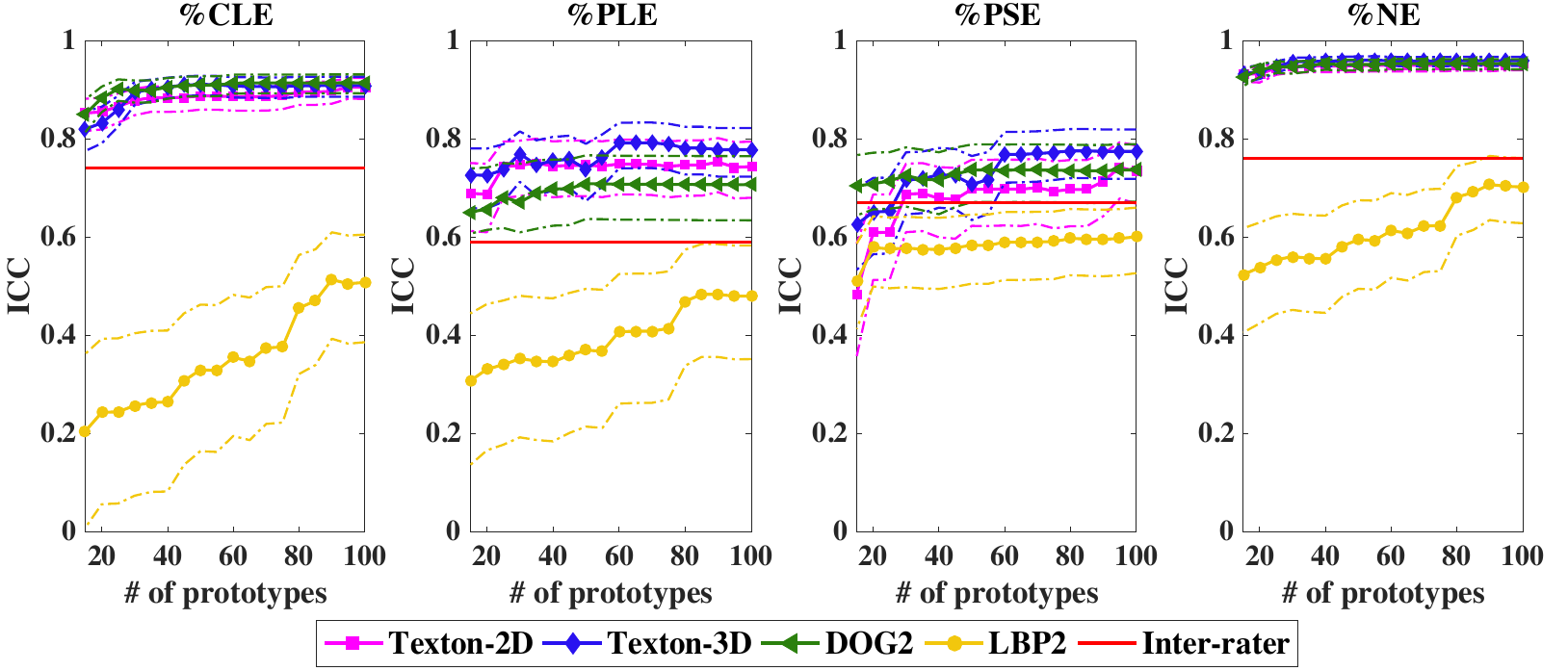}
\caption[ ] {Intraclass correlation (ICC) between predicted global labels and ground truth versus number of merged prototypes (dashed line: 95$\%$ confidence interval).}
\label{fig:proto-pruning}
\end{figure}

The quality of the predictions is evaluated using intraclass correlation (ICC) with ground truth global labels. To achieve a balance between the number of training scans (large enough to learn lung textures) and the number of test scans (large enough so that the prediction performance is not biased by extreme points), we used a 4-fold cross validation setup, with $3/4$ of scans used for training, and $1/4$ used for testing. All features were computed within 3D ROIs. Texton-based features were also extracted in 2D ROIs for comparison. We select $K=100$ as our benchmark value, from which we iteratively merge prototypes. We report the evolution of prediction capabilities as $K$ is reduced in Fig. \ref{fig:proto-pruning} (all $p-$values $<$ 0.01).

Overall, texton-based and DOG2 features give robust prediction that outperform the intra-rater agreement, while LBP2 features have poor to modest prediction capabilities. One reason might be that intensity information in LBP2 is compressed with our current feature length, while intensities improved the discriminative capability of the original LBP code in \cite{lbp2}. However, we observed that a feature length over 50 decreases the robustness and drastically increases the convergence time for unsupervised prototype clustering. This makes LBP2 less favorable in our unsupervised learning context.

The comparison of 2D versus 3D ROIs with texton-based features indicates that the richer information in 3D neighborhood is helpful for modeling emphysema subtypes, at the price of additional computational cost for feature extraction. 

Regarding the effect of prototype merging, ICC values remain steady when $K>60$ for texton-based features. Merging is capable of reducing model complexity with little sacrifice in prediction performance. For DOG2 features, the performance begins to decrease only after $K<50$. For LBP2 features, however, the performance degrades immediately after merging, which may be because the LBP2-based prototypes are not sufficiently homogeneous from the beginning.

Note that using a high number of $K$, much larger than the number of standard emphysema subtypes or than required for predictive power of these subtypes, is driven by our goal to be able to discover finer emphysema subtypes. The current arbitrary number $K=100$ will be further trimmed with an optimization metric incorporating respiratory symptoms and generalization capabilities to other datasets, which is ongoing work of our study.

\begin{figure}[t]
\includegraphics[width=12cm]{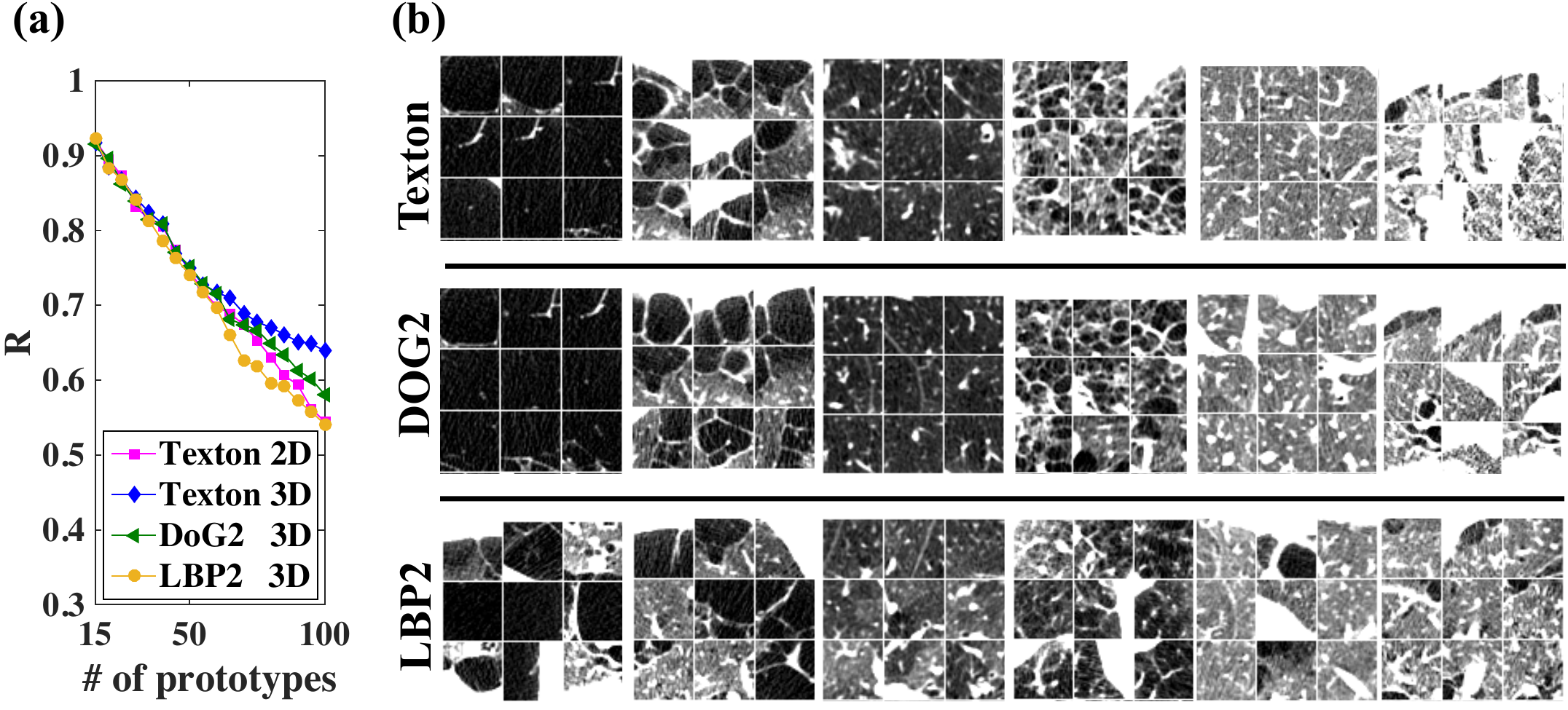}
\caption[ ] {(a) Reproducibility metric versus number of merged prototypes. (b) Examples of axial cuts from ROIs in six prototypes with three feature types. The texton-based prototypes are selected as the $1^{st}$, $5^{th}$, $20^{th}$, $40^{th}$, $80^{th}$ and $95^{th}$ benchmark prototypes. The DOG2 and LBP2-based prototypes are those having the most overlap with texton-based prototypes for ROI labeling. Window level: [-1000, -700] HU.}
\label{fig:proto}
\end{figure}

\subsection{Reproducibility of Prototypes}

Reproducibility of prototypes is measured by computing the overlap of prototype labeling with two distinct training sets (by randomly dividing the subjects into two groups), in a manner similar to \cite{resampling}. Formally, we measure:

\begin{equation}
R(L,L')=\mathrm{max}_\pi\frac{1}{K}\sum_{k=1}^{K}{\mathbb{1}(L(X_k)=\pi(L'(X_k)))}
\end{equation}
where $L$ and $L'$ are prototype labeling with two different training sets, $\mathbb{1}$ is the 0-1 loss function, $X_k$ denotes ROIs labeled with prototype $k$, and $\pi$ denotes the permutations of the $K$ prototypes using the Hungarian method \cite{resampling} for optimal matching. 

Fig. \ref{fig:proto} (a) plots $R$ versus number of prototypes. For $K<50$, reproducibility is high ($R>0.7$) for all types of features. When $K>60$, 3D texton-based prototypes are more reproducible ($R>0.6$ with $K$ as large as 100).

\begin{figure}[htbp]
\centering
\includegraphics[width=12cm]{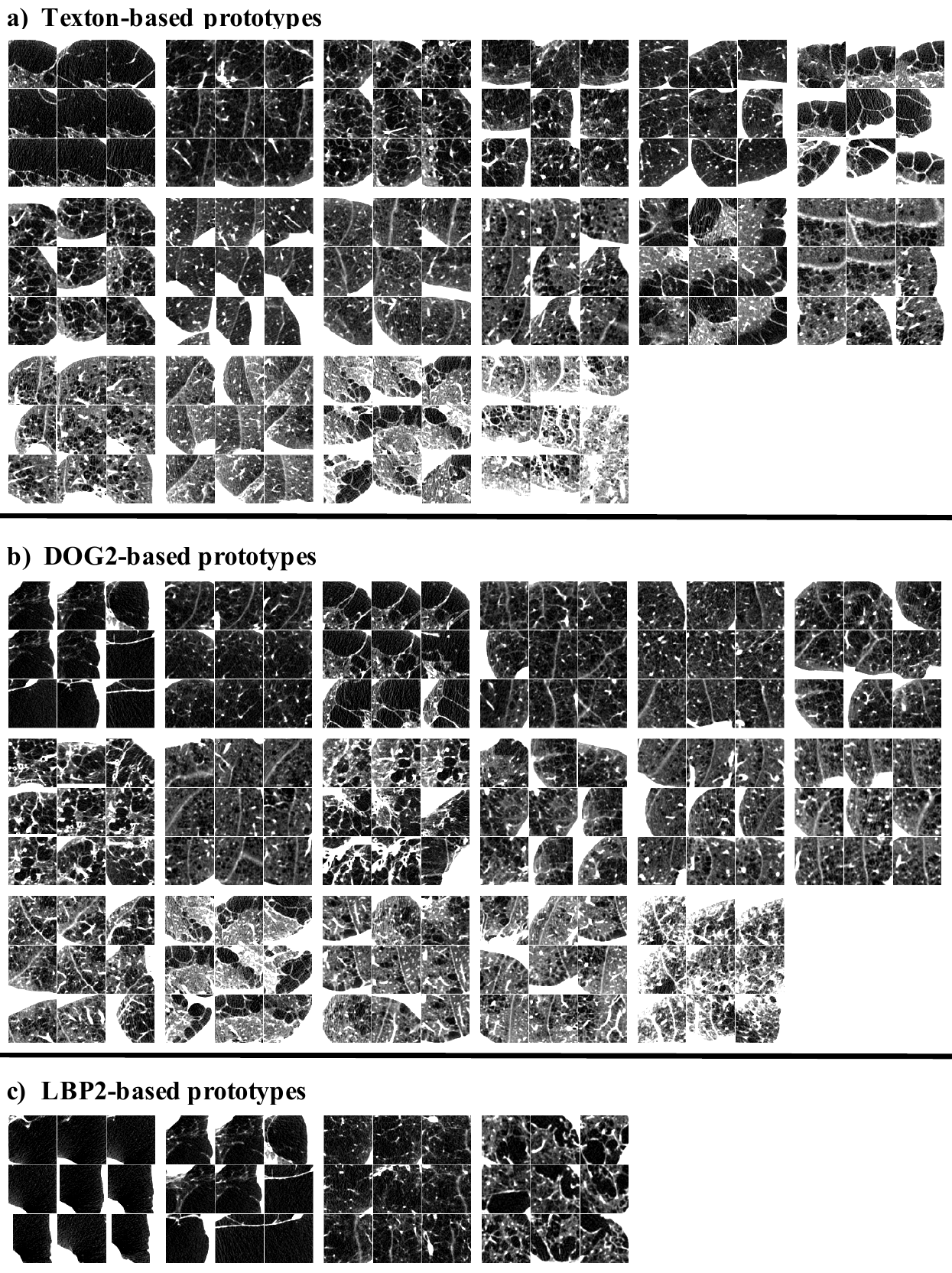} 
\caption[ ] {Axial cuts of 3D ROIs from subsets of prototypes generated with either texton-based, DOG2 or LBP2 features and that have higher occurrence in subjects with emphysema than in normals. Window level: [-1000 -700] HU.}
\label{fig:disease_like}
\end{figure}

\subsection{Visualization of Sample Prototypes}

Visual examples of prototypes generated with three different types of features using 3D ROIs are provided in Fig. \ref{fig:proto} (b). Texton and DOG2-based prototypes have high intra-class similarity and show clearly distinct lung tissue patterns, while LBP2-based prototypes have lower intra-class homogeneity, which agrees with the poorer prediction results. 

We also provide in Fig. 4 visual examples of prototypes that are likely to encode emphysematous lung tissues. 

First, subjects in the dataset were separated into two groups: \textit{disease} (visually assessed extent of emphysema [2] larger than 0) and \textit{normal} (visually assessed extent of emphysema equals to 0). 

Out of the K = 100 benchmark prototypes, we selected the ones for which occurrence within the disease population was 3 times higher than in the normal population. This lead to subsets of $n=16,17,4$ disease prototypes when using respectively texton-based, DOG2 and LBP2 features, in 3D ROIs. These subsets are illustrated in Fig. \ref{fig:disease_like} on group of 9 patches of size of $50\mathrm{mm}^3$ from random disease subjects. The large patch size (twice the length of the ROIs used for prototype generation) is used to reveal the presence of nearby lung borders.

\section{Conclusions}
In this work, we presented a novel framework to generate unsupervised lung texture prototypes that can be used to predict the overall extent of standard emphysema subtypes from a heterogeneous database of lung CT scans, using standard radiological global labels as the ground truth. We cluster unlabeled local ROIs into texture prototypes, and encode lung CT scans with prototype histograms. Labeling of ROIs is tested in 2D or 3D, and using three types of features. 

The intraclass correlations between prediction and ground truth labeling indicate that texton and DOG2 features are capable of learning homogenous prototypes and lead to very robust predictions of standard emphysema labels that outperform the inter-rater agreement, while LBP2 feature is less discriminative (at least with similar feature vector length). 

We tested model reduction via prototype merging based on inter-prototype distance and spatial co-occurrence. Results show that robust prediction can be achieved with at least $K$=60 merged prototypes for texton-based features and $K$=50 for DOG2 features. Reproducibility of texton-based prototypes is superior when $K>60$. These homogeneous and reproducible texture prototypes show potential in new interpretations of lung CT scans with finer subtyping. Since texture prototypes link image analysis-based discovery with radiological prior knowledge, and enable automated labeling of lung volumes and generation of scan signatures, they can be used for multiple tasks such as correlation with omic measures, sub-phenotyping of emphysema or image indexing and retrieval. Our future work will focus on two aspects: 1) As texton-based feature and DOG2 feature both demonstrated good capability at discovering lung texture prototypes, we would like to explore their combination to boost robustness and discovery power, which can be achieved by either feature concatenation followed by feature dimension reduction (to reduce the computational complexity, as in \cite{R1}), or post-clustering ensembling \cite{emsembling}; 2) The number of prototype $K$ will be further trimmed to find clinically significant sub-categories of emphysema, with an optimization metric incorporating clinical data and generalization capability.

\subsubsection{Acknowledgements:} Funding provided by NIH/NHLBI R01-HL121270, R01-HL077612, RC1-HL100543, R01-HL093081 and N01-HC095159 through N01-HC-95169, UL1-RR-024156 and UL1-RR-025005.

\bibliographystyle{splncs03_mod}

\end{document}